\documentclass[letterpaper]{article}
\usepackage{aaai2026}  
\usepackage{times}
\usepackage{helvet}
\usepackage{courier}
\usepackage[hyphens]{url}
\usepackage{graphicx}
\usepackage{amsmath} 
\usepackage{multirow}
\usepackage{booktabs}
\usepackage{mathrsfs}
\usepackage{amsfonts}
\usepackage{amssymb}  
\usepackage{array}   
\usepackage{booktabs} 

\usepackage{natbib}
\usepackage{caption}
\usepackage{algorithm}
\usepackage{algorithmic}
\usepackage{newfloat}
\usepackage{listings}

\DeclareCaptionStyle{ruled}{labelfont=normalfont,labelsep=colon,strut=off} % DO NOT CHANGE THIS
\floatstyle{ruled}
\newfloat{listing}{tb}{lst}{}
\floatname{listing}{Listing}
\title{PMGS: Reconstruction of Projectile Motion Across Large Spatiotemporal Spans via 3D Gaussian Splatting}

\author{
    Yijun Xu\textsuperscript{\rm 1}\equalcontrib,
    Jingrui Zhang\textsuperscript{\rm 2}\equalcontrib,
    Yuhan Chen\textsuperscript{\rm 3},
    Dingwen Wang\textsuperscript{\rm 2},
    Lei Yu\textsuperscript{\rm 4},
    Chu He\textsuperscript{\rm 1}\thanks{Corresponding author.} 
}

\affiliations{
    \textsuperscript{\rm 1}School of Electronic Information, Wuhan University\\
    \textsuperscript{\rm 2}School of Computer Science, Wuhan University\\
    \textsuperscript{\rm 3}College of Mechanical and Vehicle Engineering, Chongqing University\\
    \textsuperscript{\rm 4}School of Artificial Intelligence, Wuhan University\\

    \{xyj021021, zjr233, chuhe\}@whu.edu.cn
}

\begin{document}
\maketitle
\begin{abstract}

Modeling complex rigid motion across large spatiotemporal spans remains an unresolved challenge in dynamic reconstruction. Existing paradigms are mainly confined to short-term, small-scale deformation and offer limited consideration for physical consistency. This study proposes \textbf{PMGS}, focusing on reconstructing \textbf{P}rojectile \textbf{M}otion via 3D \textbf{G}aussian \textbf{S}platting. The workflow comprises two stages: 1) Target Modeling: achieving object-centralized reconstruction through dynamic scene decomposition and an improved point density control; 2) Motion Recovery: restoring full motion sequences by learning per-frame SE-3 poses. We introduce an acceleration consistency constraint to bridge Newtonian mechanics and pose estimation, and design a dynamic simulated annealing strategy that adaptively schedules learning rates based on motion states. Furthermore, we devise a Kalman fusion scheme to optimize error accumulation from multi-source observations to mitigate disturbances. Experiments show PMGS’s superior performance in reconstructing high-speed nonlinear rigid motion compared to mainstream dynamic methods.
\end{abstract}
\begin{links}
\link{Code}{https://github.com/X-Probiotics/PMGS}
\end{links}

\section{Introduction}

Dynamic reconstruction has become an engine driving modern film animation, game interaction, and virtual reality. The rise of neural rendering has pushed reconstruction fidelity to unprecedented heights, enabling the depiction of highly challenging deformations, such as subtle tremors in biological tissues \cite{Endo}. Meanwhile, breakthroughs in generative methods \cite{Dream4DGS, Dreamscene4D} have achieved controllable synthesis of dynamic scenes with diverse artistic styles, vastly expanding the boundary of imagination in visual expression.

However, when returning to a task governed by the essential laws of the physical world—\textbf{the reconstruction of complex rigid motions over large spatiotemporal spans}—there lies a challenging issue that remains insufficiently explored.

Most existing modeling paradigms are designed for small range non-rigid motion over short time spans \cite{DynaNeRF2, DyNeRF, DeformGS, 4DGS}. These approaches rely on deformation fields with complex temporal regularization to achieve spatiotemporal alignment. However, discrete sampling struggles to capture large nonlinear motion (e.g., accelerated rotation), often resulting in noticeable artifacts and trajectory fragmentation when modeling high-speed rigid motion \cite{4DGS}. Additionally, current 4D datasets exhibit similar limitations: in most scenarios, moving objects occupy only a small portion of the scene and exhibit minimal movement. Consequently, many methods prioritize optimizing appearance similarity metrics rather than addressing the core challenge of accurate motion recovery.

\begin{figure}[t]
\centering
\includegraphics[width=\linewidth]{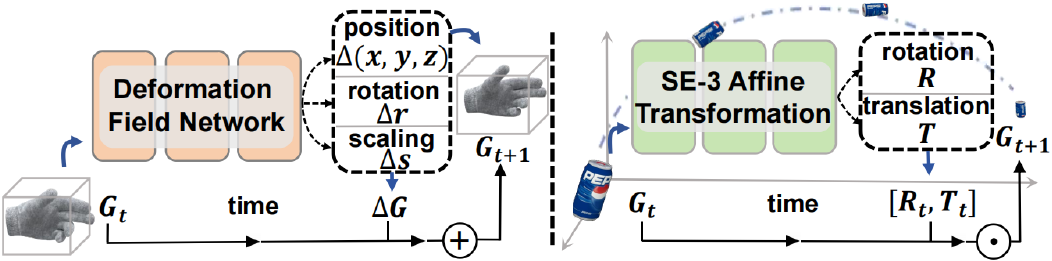}
\caption{Left: Current paradigms focus on small-scale deformation reconstruction.  
Right: PMGS explores complex rigid motion modeling across large spatiotemporal spans.}
\label{Intro}
\end{figure} 

The absence of physical consistency poses another major challenge. Many works focus on optimizing appearance similarity while neglecting the pervasive physical constraints in the real world. However, when modeling complex motion across large spatiotemporal spans, relying solely on photometric supervision fails to constrain the vast 3D solution space effectively. For NeRF-based methods \cite{NeRFPlayer,DyNeRF, DynaNeRF3}, the inherent characteristics of implicit representations encode motion cues within black-box network weights, making physical interactions difficult to model and unintuitive. The introduction of 3D Gaussian Splatting (3DGS) \cite{3DGS} has partially addressed this by combining explicit point-based representations with neural rendering, thereby providing a structured carrier for scene understanding. Consequently, some GS-based approaches \cite{Endo, CF, PEGAUS} incorporate off-the-shelf models to enhance physical realism. We argue that this model-stacking strategy introduces compounded uncertainties by layering new approximations over existing ones. More critically, it bypasses the exploration of fundamental physical principles, ultimately reducing the reconstruction task to uninterpretable compensatory computations.

To tackle these issues, we selected a representative task for focused solutions: \textbf{Projectile motion}. As a fundamental concept in mechanics, it describes an object launched with an initial velocity, following parabolic motion under gravity. The study extends from free-fall to complex situations with self-rotation. This phenomenon is not only ubiquitous in the real world (e.g., falling objects from heights, throwing events in sports, military ballistic trajectories), but also encompasses the typical challenges of rigid motion, including variable speed, large span, and compound motion (translation and rotation). Consequently, this archetypal paradigm can be generalized to any dynamic scenario within constant force fields in nature, demonstrating universal applicability.

In this paper, we focus on the modeling of complex rigid motions with large spatiotemporal spans, and propose \textbf{PMGS}, a framework that reconstructs \textbf{P}rojectile \textbf{M}otion via 3D \textbf{G}aussian \textbf{S}platting. By leveraging temporal continuity in video sequences and explicit Gaussian representation, we estimate per-frame SE(3) affine transformation throughout the motion which is fundamentally unattainable with NeRF-based methods. Guided by first principles of physics, we introduce an acceleration constraint for motion recovery. Additionally, to alleviate issues of learning oscillations and trajectory fractures caused by fixed training paradigms, we develop a Dynamic Simulated Annealing (DSA) strategy that adaptively schedules the training process based on real-time velocity and displacement variations. Departing from conventional methods that treat physical constraints merely as regularization components, we propose a Kalman filter-based optimal estimation scheme to fuse cross-modal observations, thereby minimizing potential disturbances.

Overall, our main contributions can be outlined as: 
\begin{itemize}

\item We propose PMGS, a framework integrating target modeling and motion recovery to reconstruct projectile motion. Particularly, we contribute a dataset designed for complex rigid motion modeling at large spatiotemporal scales—an area notably underserved by existing datasets.

\item We introduce an acceleration consistency constraint connecting Newtonian mechanics to pose estimation. Our DSA strategy dynamically adapts training via velocity-displacement motion states, addressing fixed paradigms' nonlinear motion limitations.

\item We design a Kalman fusion scheme to dynamically optimize multi-source observations with adaptive weighting, minimizing error accumulation caused by real-world disturbances or training oscillations.

\end{itemize}

\begin{figure*}[t]
    \centering
    \includegraphics[width=\textwidth]{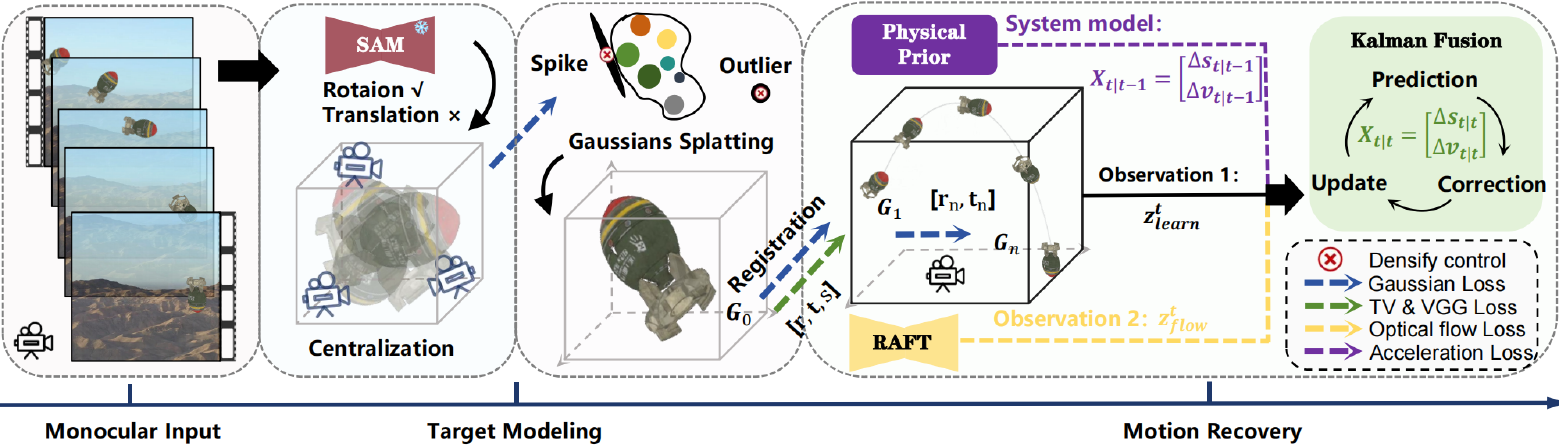} 
    \caption{Overview of PMGS. We first segment the target, then decompose the motion through centralization to transform the dynamic scene into a static one. For modeling, we learn a set of Gaussian kernels and align them at the original scale with a set of learnable affine transformations. In motion recovery, we estimate the target's SE(3) transformation frame by frame, and comprehensively improve tracking accuracy by integrating physics-enhanced strategies.}
    \label{fig:doublecolumnimage}
\end{figure*}

\section{Related Work}

\subsubsection{Dynamic Neural Rendering.}
Current dynamic neural rendering includes: (1) Point deformation fields \cite{DynaNeRF2} map sampled points at each timestep to a static canonical space. (2) Time-aware Volume Rendering \cite{DyNeRF, NeRFPlayer} employs factorized voxels to independently compute features for each point. (3) The Gaussian deformation field \cite{DeformGS, MotionGS, S4d} transforms 3D Gaussians to their target position at specific timesteps. The critical issues lie in that the deformation field struggles to model large inter-frame displacements that commonly occur in real-world scenarios, which can be further validated by experiments on Panoptic Studio dataset \cite{Sportsdatasest} conducted by 4DGS \cite{4DGS}. Moreover, due to the lack of physical consistency constraints, the network tends to focus on fitting appearance similarity while neglecting structural accuracy at the geometric level. As a typical example, when there are large areas of nearly uniform color, Gaussian bodies will float chaotically within regions of similar color \cite{Dynamic3DGS}.

\subsubsection{Enhancing GS with Physical Assistance.}
Introducing physical assistance to enhance 3DGS has been widely applied. \cite{CF, Endo, FSGS} utilize off-the-shelf monocular depth estimation models \cite{DPT} for scene initialization or sparse reconstruction. \cite{Spring} devised a Spring-Mass model to simulate the elastic objects' falling and collision. \cite{PEGAUS} is designed for 3D dataset generation which integrates the PyBullet engine to emulate the placement of objects and their dynamic processes. \cite{Dreamscene4D, SAGS} enhances pose estimation by ensuring scale consistency and local rigidity. Essentially, current approaches use rough approximations from external physics models to fix ill-posed inverse problems in vision systems. Nevertheless, pre-trained engines lose confidence sharply in unusual scenarios, and their black-box makes errors untraceable.

\section{Method}
PMGS takes a monocular video as input to reconstruct the target and achieve full-sequence projectile motion recovery. For modeling appearance and geometry, we equivalently convert dynamic scenes into static through motion decomposition, and combine with the improved point density control strategy to enhance the geometric accuracy. During motion recovery, we leverage video temporal continuity and the explicit Gaussian representation to estimate per-frame SE(3) transformations. To ensure physically consistent pose learning, we introduce an acceleration consistency constraint that establishes direct connections between pose estimation and Newtonian dynamics priors. Finally, we design a Kalman fusion strategy to optimize the error accumulation from multi-modal observation sources.

\subsection{Target Modeling} 
To model the target's appearance and geometry, we reformulate the dynamic scene as a static object-centralized scene.

\textbf{Centralization.} We firstly use the pre-trained SAM \cite{SAM} to separate the background and the dynamic target. Then, by establishing object-centralized normalized coordinates \cite{Dreamscene4D}, we decompose the complex projectile motion: the object's autorotation is disentangled into pseudo multi-view observations, while translational displacements are eliminated.

\textbf{Reconstruction.} Following the 3DGS \cite{3DGS}, we use a set of Gaussian kernels to explicitly reconstruct the appearance and geometry of the object.

Notably, geometric quality directly affects motion recovery. Therefore, the structural flaws of Gaussian representation cannot be ignored: (1) Oversize Gaussians. They frequently occur at boundaries to incorrectly fit the viewpoint variations, inducing jagged edges and spiky artifacts. (2) Discretized Gaussians. They appear distant from the target, distorting centroid computation and introducing systematic errors in motion estimation. To address these, we propose an improved point density control strategy: (1) Axial constraint: Hard pruning of Gaussians with excessive axial lengths; (2) Outlier removal: Gaussians whose distance to their nearest neighbors exceeds the average value are removed. This can be mathematically expressed as:
\begin{equation}
\mathcal{G}_{\text{pruned}} = 
\left\{ 
g_i \in \mathcal{G} \,\middle|\,
\begin{aligned}
&\max\big(\mathrm{eig}(\Sigma_i)\big) \leq \tau_L \quad \cap \\
&\min_{g_j \in \mathcal{G}} \|\mu_i - \mu_j\|_2 \leq \tau_D \cdot D_{avg}
\end{aligned}
\right\}
\end{equation}
where $\Sigma_i$ and $\mu_i$ represent the covariance matrix and position of a Gaussian $g_i$ respectively. For axial constraint, Gaussians with length of principal axis $\max\big(\mathrm{eig}(\Sigma_i)\big) > \tau_L$ are discarded, where $\tau_L$ is a predefined threshold controlling shape anisotropy. Besides, an outlier is defined as a Gaussian whose minimum distance to others is much larger than the average distance $D_{avg} = \frac{1}{|\mathcal{G}|^2} \sum_{m,n} \|\mu_m - \mu_n\|_2$, where $\tau_D$ is a distance-based filtering factor.

During the optimization, the loss $\mathcal{L}_{GS}$ is used to quantify the difference between the real image $\hat{I}$ and rendered image $I$:
\begin{equation}
\mathcal{L}_{GS} = (1-\lambda)\lVert{\hat{I}-I}\rVert_{1} + \lambda\mathcal{L}_{D-SSIM}\label{Equation4}
\end{equation}

Up to this point, we obtain a Gaussian field $G_0$ at the centralized scale.

\textbf{Registration.} Finally, we register the static Gaussian $G_0$ at the centralized scale to the original dynamic scene, thus facilitating subsequent motion recovery. This can be achieved via a set of learnable affine transformations $T_{reg} = [\mathbf{r, t}, s]$. Then the aligned Gaussian field $G_1$ can be represented as $T_{reg}\odot G_0$. In addition to $\mathcal{L}_{GS}$, we further apply VGG loss $\mathcal{L}_{VGG}$ and grid-based total variation loss $\mathcal{L}_{TV}$ \cite{Spring, 4DGS} to achieve a better awareness of accurate registration. The total loss can be formulated as:
\begin{equation}
\mathcal{L}_{Align} = \lambda_{1}\mathcal{L}_{GS}+ \lambda_{2}\mathcal{L}_{VGG} + \lambda_{3}\mathcal{L}_{TV}\label{Equation7}
\end{equation}

The optimization target is to minimize $\mathcal{L}_{Align}$ between the rendered image of $G_1$ and the first frame $\hat{I}_0$ of original dynamic scene:
\begin{equation}
T_{reg} = \mathrm{arg}\underset{T_{reg}}{\mathrm{min}}\mathcal{L}_{Align}(\mathcal{R}(T_{reg}\odot G_0),\hat{I}_0)\label{Equation6}
\end{equation}

Finally, we obtain this set of Gaussians $G_1$, which can well represent the appearance and geometry of the object.

\begin{figure}[t]
    \centering
    \includegraphics[width=\linewidth]{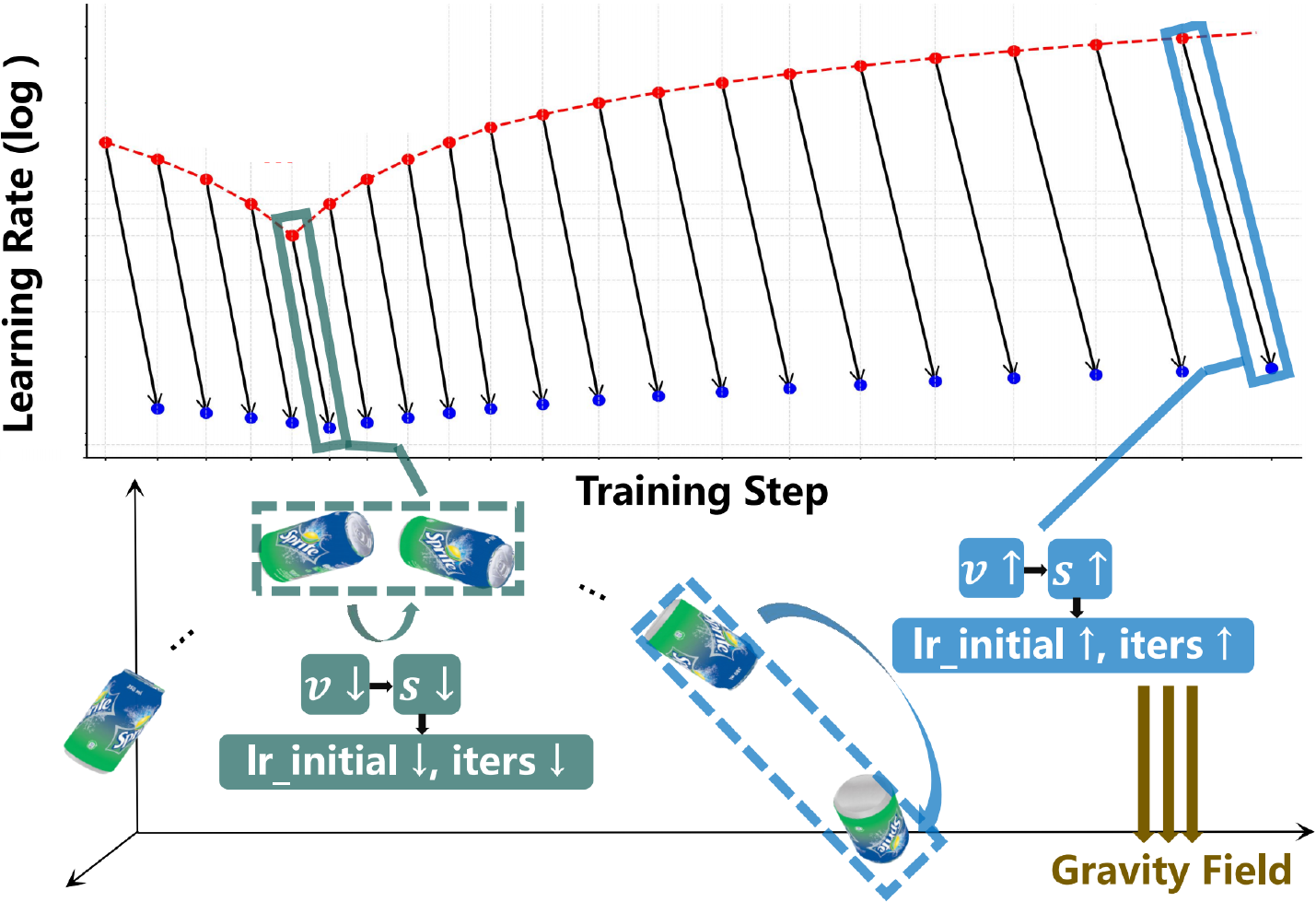} 
    \caption{DSA strategy. Right: An object accelerates in a constant gravity field, with different velocities $v$ and displacements $s$ corresponding to each timestamp. Left: Red curve represents the initial learning rate. Blue points show the final learning rate after exponential decay.}
    \label{DSA}
\end{figure} 

\subsection{Physics-Enhanced Motion Recovery}

During the dynamic phase, we estimate 6DoF pose changes of the object in total $(n+1)$ frames, which correspond to a sequence of temporal SE-3 transformations $T_n = [\mathbf{r}_n, \mathbf{t}_n]$. We introduce an acceleration consistency constraint to establish a direct connection between pose learning and physical priors. Additionally, we incorporate optical flow smoothing and a DSA training strategy to improve training stability.

\textbf{Acceleration consistency constraint.} The object moves in a constant force field, therefore its acceleration remains constant. Based on this, we first calculate the object's center of mass $\sigma$. Since the Gaussians have been processed isotropically, the spatial coordinates of $\sigma$ can be computed as:
\begin{equation}
\sigma = \frac{\sum_{i=1}^NR_i^3\mu_i}{\sum_{i=1}^NR_i^3}
\label{Equation8}
\end{equation}
where $N$ represents the total number of Gaussian kernels, $R$ is the radius of each kernel, and $\mu_i = [x_i,y_i,z_i]$ denotes the spatial position of the kernel. In turn, the acceleration of the object at the current moment $t$ can be calculated as:
\begin{equation}
{a}^{t}=\frac{[\sigma^{(t+\Delta{t})}-\sigma^t]-[\sigma^t-\sigma^{(t-\Delta{t})}]}{(\Delta t)^2}\label{Equation9}
\end{equation}
where $\Delta{t}$ denotes the time interval between two adjacent frames. Subsequently, the acceleration $\textbf{a}^{t}$ can be decomposed into components parallel and orthogonal to the gravity direction $\textbf{g}$:
\begin{equation}
\begin{aligned}
\mathbf{a}^{t}_{\parallel} &= (\mathbf{a}^{t} \cdot \mathbf{g})\, \mathbf{g}, \\
\mathbf{a}^{t}_{\perp} &= \mathbf{a}^{t} - (\mathbf{a}^{t} \cdot \mathbf{g})\, \mathbf{g}
\end{aligned}
\label{acceleration}
\end{equation}

According to Newtonian dynamics, there should be a constant acceleration along the gravity direction, while the components orthogonal to it should remain zero:
\begin{equation}
\mathcal{L}_{Acc} = \lVert{\mathbf{a}^{t}_{\parallel} + (\mathbf{a}_{\perp}^{(t+\Delta{t})} 
 - \mathbf{a}_{\perp}^{t})}\rVert_{2}^{2}
\label{lossacc}
\end{equation}

Building upon Eq.\ref{lossacc}, we are able to effectively regularize the learning of translation components across consecutive frames, even under scale ambiguity, thereby ensuring the physical consistency of motion recovery without requiring absolute values such as gravitational acceleration.

\textbf{Optical flow smoothing.}
We introduce optical flow smoothing computed based on \cite{Optical} for emphasizing attention to motion variations in long-term tracking. In addition to evaluating the similarity between the real optical flow $\hat{F}$ and rendered $F$, we compute the gradients of the horizontal and vertical components of the optical flow field, applying a smoothness penalty with different weights for corresponding regions:
\begin{equation}
\begin{aligned}
\mathcal{L}_{Smooth}& = \lambda_1[\frac{1}{N}\sum_{i=1}^N(\Delta F_{i,x}\cdot\exp(-\frac{\Delta I_{i,x}}{10}) + \\
&\Delta F_{i,y}\cdot\exp(-\frac{\Delta I_{i,y}}{10}))] + \lambda_2\lVert{\hat{F}-F}\rVert_{1}
\label{Equation11}
\end{aligned}
\end{equation}
where $\Delta F$ and $\Delta I$ denote the gradients of the optical flow field and the rendered image respectively.

\textbf{Dynamic simulated annealing.} As shown in Figure \ref{DSA}, a fixed $lr_{init}$ struggles to adapt to varying object's velocities: excessively large $lr_{init}$ induces training oscillations during low-speed phases, whereas small $lr_{init}$ hinders convergence when tracking high-speed targets. To address this velocity-displacement coupling effect, we dynamically schedule the $lr_{init}$ according to real-time displacement.

Specifically, for a moving object with an initial velocity $v_0$ and constant acceleration $a$, the displacement during the interval from timestamp $t$ to $(t+\Delta t)$ can be derived as:
\begin{equation}
\Delta s_t = (a\Delta t)t+[\frac{1}{2}a(\Delta t)^2+v_0\Delta t]
\label{Equation12}
\end{equation}
As the initial velocity $v_0$, acceleration $a$ and the time interval $\Delta t$ are constant values, Eq.\ref{Equation12} establishes the linear correlation between object's displacement $\Delta s$
and timestamp $t$, i.e., 
$\Delta s \propto t$. This temporal-displacement relationship therefore requires proportional scaling of the $lr_{init}$ with displacement. Therefore, we implement scheduling where the learned displacement from the preceding timestep governs the subsequent $lr_{init}$, as visualized in Figure \ref{DSA}.

As the iteration progresses, the Gaussians gradually approach the target spatial position. Correspondingly, the learning rate needs to be reduced so that the step gradually shrinks for finetuning. We utilize the commonly used exponential decay to achieve this. In addition, for timestamps with large displacements, we appropriately extend the number of iterations to better find the optimal solution.

In summary, we follow the DSA strategy for frame-by-frame tracking of the pose $T_n$ during motion recovery, and the total optimization is to minimize a composite of losses:
\begin{equation}
\begin{aligned}
T_{n}& = \mathrm{arg}\underset{T_{n}}{\mathrm{min}}(\lambda_1\mathcal{L}_{GS}(\mathcal{R}(T_{n}\odot G_n),\hat{I}_{n+1})+\\
&\lambda_2\mathcal{L}_{Acc} +
\lambda_3\mathcal{L}_{Smooth}({F}(T_{n}\odot G_n),\hat{F}_{n+1})\label{Equation14}\\
\end{aligned}
\end{equation}

\begin{figure*}[t]
    \centering
    \includegraphics[width=\textwidth]{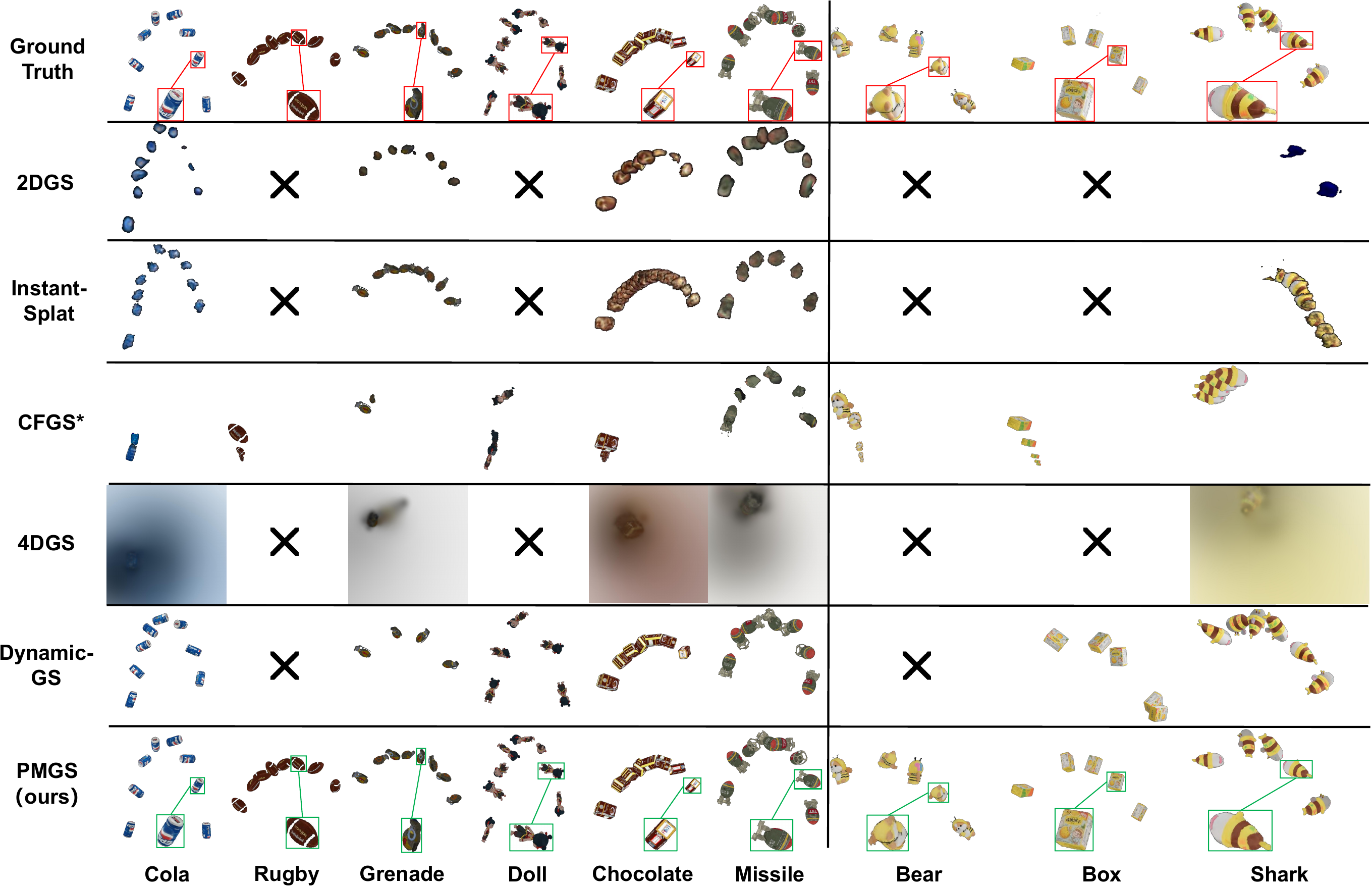} 
    \caption{Qualitative comparison on both synthetic and real datasets. PMGS generalizes well across different scenarios and accurately reconstructs full-sequence motion. The displayed results are uniformly sampled, for complete and coherent motion recovery, please refer to the video on the project page.}
    \label{comparison}
\end{figure*}

\subsection{Cross-Modal Kalman Fusion}
During the aforementioned stage, there exist different sources of cross-modal observations. Some works \cite{MotionAware} also incorporate optical flow or other priors as auxiliary inputs. However, they often fail to account for inaccurate optical flow estimations caused by motion-blurred and texture-degraded regions. This oversight, in turn, introduces negative effects during optimization. Differently, we design a data fusion strategy based on Kalman filter \cite{Kalman} to dynamically balance the weights of observation sources, suppress cumulative error, and update the optimal estimate in real-time.

\noindent
\textbf{Step1-Definition: }The basic elements in Kalman fusion namely the system prediction model and observations. When applied to our motion recovery process, these components correspond to the following:

\begin{itemize}
\item System model: Displacement prediction based on the acceleration consistency constraint (Eq.8). 

\item  Observation 1: The displacement calculated via back-projection from the inter-frame optical flow field $z_{flow}^{t}$.

\item  Observation 2: The displacement of the current frame actually learned by the network $z_{learn}^{t}$.
\end{itemize}

\noindent
\textbf{Step2-Prediction: }In the prediction step, we use the system dynamic model and previous state estimation to predict the current state, which can be formulated as: 
\begin{equation}
\begin{aligned}
& \textbf{X}_{t|t-1} = \begin{bmatrix}
  \Delta s_{t|t-1} \\
  v_{t|t-1}
\end{bmatrix} = \textbf{F}\textbf{X}_{t-1|t-1} + \textbf{B}a_t + \textbf{w}_{t}, \\
& \textbf{F} = \begin{bmatrix}
  1 & \Delta t \\
  0 & 1 
  \end{bmatrix}, \textbf{B} = \begin{bmatrix}
   \frac{1}{2}(\Delta t)^2\\
  \Delta t \end{bmatrix}, \textbf{Q} =\begin{bmatrix}
  \sigma_{\Delta s}^{2} & 0 \\
  0 & \sigma_{v}^{2} 
  \end{bmatrix}
\label{predict}
\end{aligned}
\end{equation}
where $\textbf{X}_{t|t-1}$ is the prediction based on the optimal estimate at the previous timestamp, \textbf{F} is the state transition matrix, \textbf{B} is the control input matrix, and $a_t$ is the control input calculated from Eq.6, which approximates to $a_{t-1}$ as the system model defined. The prediction covariance can then be calculated as:
\begin{equation}
\begin{aligned}
& \textbf{P}_{t|t-1} = \textbf{F}\textbf{P}_{t-1|t-1} \textbf{F}^{T} + \textbf{Q}
\label{P}
\end{aligned}
\end{equation}
where $\textbf{P}_{t-1|t-1}$ is the posterior covariance of the previous state.

\begin{table}[t]
\centering
\renewcommand{\arraystretch}{1}
\setlength{\tabcolsep}{3pt}

\resizebox{0.48\textwidth}{!}{
\begin{tabular}{l c c c c c c}
\hline
Datasets & Source & Type & Phy. & Mono. \\
\hline
LaSOT \cite{lasot}  & Real   & Rigid  &  $\times$ & $\times$   \\
D-NeRF \cite{DynaNeRF2}     & Syn.   & Def.  &  $\times$ & $\checkmark$  \\
HyperNeRF \cite{DynaNeRF3}   &  Real & Def. & $\times$ & $\checkmark$\\
Neural3D  \cite{DyNeRF} &Syn. & Def. & $\times$ & $\times$ \\
PEGASUS \cite{PEGAUS}   & Syn.      &  Rigid       & PyBullet   & $\checkmark$    \\
SpringGS \cite{Spring}   & 6S + 3R & Def. &  Hooke's Law &  $\times$     \\
PMGS   & 6S + 3R  & Rigid & Newton's Law & $\checkmark$ \\
\hline
\end{tabular}
    }
\caption{Datasets comparison (Phy.-Whether physical prior is introduced; Mono.-Whether monocular; Syn.-Synthetic; Def.-Deformable).}
\label{datsests}
\end{table}

\begin{table*}[t]
    \centering

    \renewcommand{\arraystretch}{1}
    \setlength{\tabcolsep}{2pt}
    \resizebox{1\textwidth}{!}{
    \begin{tabular}{l|*{5}{ccc|}cccc}
        \toprule
        & \multicolumn{3}{c|}{Grenade} 
        & \multicolumn{3}{c|}{Missile} 
        & \multicolumn{3}{c|}{Cola} 
        & \multicolumn{3}{c|}{Doll}
        & \multicolumn{3}{c|}{Chocolate} 
        & \multicolumn{3}{c}{Rugby} 
\\
        
        \cmidrule(lr){2-4}
        \cmidrule(lr){5-7} 
        \cmidrule(lr){8-10} 
        \cmidrule(lr){11-13} 
        \cmidrule(lr){14-16} 
        \cmidrule(lr){17-19}

        & PSNR$\uparrow$ & SSIM$\uparrow$ & LPIPS$\downarrow$ 
        & PSNR$\uparrow$ & SSIM$\uparrow$ & LPIPS$\downarrow$
        & PSNR$\uparrow$ & SSIM$\uparrow$ & LPIPS$\downarrow$ 
        & PSNR$\uparrow$ & SSIM$\uparrow$ & LPIPS$\downarrow$ 
        & PSNR$\uparrow$ & SSIM$\uparrow$ & LPIPS$\downarrow$ 
        & PSNR$\uparrow$ & SSIM$\uparrow$ & LPIPS$\downarrow$ 
 \\
        \midrule

        2DGS
        &24.82  &0.895  & 0.153
        &17.42  &0.714  &  0.325
        &15.97  &0.847  &  0.195
        &17.99  &0.792 & 0.218
        &15.19  &0.758 &  0.275
        &21.56  &0.830 &  0.185
     \\
        
        In-Splat
        &25.44 &0.900  & 0.139
        &17.58  &0.724  &   0.300
        &17.03   &0.852   &  0.175
        &18.10  &0.797 &  0.221
        &16.10 &0.761 & 0.255
        &24.21 &0.877  & 0.165
        
     \\
        
        CFGS*
        & 21.82 & 0.873 & 0.231
        & 13.70 & 0.675 & 0.259
        & 14.47 & 0.845 &  0.284
        & 18.00 & 0.788 & 0.307
        & 11.93  & 0.739  &  0.317
        &21.37 &0.822 & 0.196
      \\

        4DGS
        & 20.52  & 0.907  & 0.112
        & 22.60  & 0.829  & 0.082
        & 20.25  & 0.859   & 0.154
        & 23.92  & 0.947  & 0.063
        & 17.91  & 0.934   & 0.182
        &Failed  &Failed   & Failed 
        
     \\
        
        Mo-GS
        & 22.64  & 0.876  & 0.166
        & 19.14   &  0.857 & 0.156
        & 15.39  & 0.823 & 0.217
        & 22.70  &  0.854  & 0.119
        & 16.99 & 0.902  & 0.115
         & 17.37  & 0.855  & 0.151
   \\

        Dy-GS 
        &  \textbf{36.32} & \textbf{0.971}  &   \textbf{0.064}
        & 26.37  & 0.906  & 0.032 
        & 28.44  & 0.939  & 0.026
        & 31.81  & 0.942  & 0.050  &\textbf{25.36}   & 0.897  &\textbf{0.018}
        &Failed   &Failed  &Failed  \\
        
        PMGS
        & 33.58 & 0.967 & 0.011
        & \textbf{28.87} & \textbf{0.915} & \textbf{0.025}
        & \textbf{26.50} & \textbf{0.940} & \textbf{0.014}
        & \textbf{34.38} & \textbf{0.959} & \textbf{0.014}
        & 25.01 & \textbf{0.924} & 0.020
        & \textbf{33.72} & \textbf{0.954}  & \textbf{0.012} 
        
        \\
        
        \bottomrule
    \end{tabular}
    }
        \caption{Comparison of video reconstruction (Synthetic data).}
    \label{2}
\end{table*}

\begin{table}[t]
    \centering

    \renewcommand{\arraystretch}{1}
    \setlength{\tabcolsep}{1pt}
    \resizebox{\columnwidth}{!}{
    \begin{tabular}{l|*{2}{ccc|}cccc}
        \toprule
        & \multicolumn{3}{c|}{Box}
        & \multicolumn{3}{c|}{Shark} 
        & \multicolumn{3}{c}{Bear} \\
        
        \cmidrule(lr){2-4}
        \cmidrule(lr){5-7} 
        \cmidrule(lr){8-10} 
        
        & PSNR$\uparrow$ & SSIM$\uparrow$ & LPIPS$\downarrow$ 
        & PSNR$\uparrow$ & SSIM$\uparrow$ & LPIPS$\downarrow$
        & PSNR$\uparrow$ & SSIM$\uparrow$ & LPIPS$\downarrow$  \\
        \midrule

        2DGS
        
        &19.72  &0.973 & 0.172
        &11.57  &0.794 &  0.101
        &11.78  &0.823 & 0.201\\
        
        In-Splat

        &20.48  &0.977 & 0.152
        &12.88 &0.810 & 0.097
        &11.79 &0.823  & 0.090 \\
        
        CFGS*

        &17.19 & 0.946 & 0.162
        &11.32  & 0.742  &  0.271
        &11.54 &0.772 &   0.212 \\

        4DGS

        & 20.19  & 0.874   & 0.131
        & 15.32  & 0.709  & 0.302
        
        &Failed   &Failed  & Failed \\
        
        Mo-GS

        & 15.92 & 0.512  & 0.464
        & 10.38 & 0.654  &  0.353
        & 13.99  & 0.887  & 0.231\\

        Dy-GS 

        & 18.97  & 0.704  &   0.152
        & 24.92  & 0.952  &   0.085 
        & Failed   & Failed  & Failed \\
        
        PMGS

        & \textbf{30.45} & \textbf{0.974}  & \textbf{0.019}
        & \textbf{25.07} & \textbf{0.952}  & \textbf{0.094}
        & \textbf{26.64} & \textbf{0.959}  & \textbf{0.059}
        \\
        
        \bottomrule
    \end{tabular}
    }
        \caption{Comparison of video reconstruction (Real data).}
    \label{3}
\end{table}

\begin{table*}[t]
    \centering

    \renewcommand{\arraystretch}{1}
    \setlength{\tabcolsep}{3pt}
    \resizebox{1\textwidth}{!}{
    \begin{tabular}{l|*{5}{ccc|}cccc}
        \toprule
        & \multicolumn{3}{c|}{Grenade} 
        & \multicolumn{3}{c|}{Missile} 
        & \multicolumn{3}{c|}{Cola} 
        & \multicolumn{3}{c|}{Doll}
        & \multicolumn{3}{c|}{Chocolate} 
        & \multicolumn{3}{c}{Rugby} \\
        
        \cmidrule(lr){2-4}
        \cmidrule(lr){5-7} 
        \cmidrule(lr){8-10} 
        \cmidrule(lr){11-13} 
        \cmidrule(lr){14-16} 
        \cmidrule(lr){17-19}
        
        & IoU$\uparrow$ & ATE$\downarrow$ & RMSE$\downarrow$ 
        & IoU$\uparrow$ & ATE$\downarrow$ & RMSE$\downarrow$ 
        & IoU$\uparrow$ & ATE$\downarrow$ & RMSE$\downarrow$ 
        & IoU$\uparrow$ & ATE$\downarrow$ & RMSE$\downarrow$ 
        & IoU$\uparrow$ & ATE$\downarrow$ & RMSE$\downarrow$ 
        & IoU$\uparrow$ & ATE$\downarrow$ & RMSE$\downarrow$                        
        \\
        
        \midrule

        CFGS*
        & 0.249 & 0.550 & 0.602
        & 0.203 & 0.129 & 0.218
        & 0.154 & 0.533 &  0.563
        & 0.310 & 0.358 & 0.459
        & 0.318  & 0.574  & 0.610
        & 0.094 & 0.562 & 0.604
         \\

        4DGS
        & 0.359  & 0.661  & 0.703
        & 0.608  & 0.194  & 0.248
        & 0.354  & 0.630  & 0.661
        & 0.401  & 0.563  & 0.621
        & 0.287  & 0.668  & 0.687
         &Failed   &Failed   & Failed 

         \\
        
        Mo-GS
        & 0.288  & 0.474  & 0.505
        & 0.541   &  0.655 & 0.704
        & 0.423  & 0.511 & 0.551
        & 0.501  &  0.505  & 0.559
        & 0.549 & 0.501  & 0.522
          & 0.249 & 0.643  & 0.668

     \\

        Dy-GS 
        & 0.548  & 0.227  &   0.269
        & 0.910  & 0.118  & 0.156  
        & 0.898  & 0.182  & 0.226
        & 0.663  & 0.311  & 0.363  
        &\textbf{0.995}   & \textbf{0.070}  &\textbf{0.090}
        &Failed   &Failed  & Failed 

        \\
        
        PMGS
        & \textbf{0.998} & \textbf{0.168} & \textbf{0.254}
        & \textbf{0.995} & \textbf{0.091} & \textbf{0.130}
        & \textbf{0.997} & \textbf{0.069} & \textbf{0.080}
        & \textbf{0.993} & \textbf{0.092} & \textbf{0.152}
        & 0.987 & 0.168 & 0.214
        & \textbf{0.952} & \textbf{0.021}  & \textbf{0.148}

        \\
        
        \bottomrule
    \end{tabular}
    }
   \caption{Comparison of motion recovery (Synthetic data).}
       \label{4}
\end{table*}

\begin{table}[t]
    \centering
    \renewcommand{\arraystretch}{1}
    \setlength{\tabcolsep}{1.8pt}
    \resizebox{\columnwidth}{!}
    {
    \begin{tabular}{l|*{2}{ccc|}cccc}
        \toprule

        & \multicolumn{3}{c|}{Box}
        & \multicolumn{3}{c|}{Shark} 
        & \multicolumn{3}{c}{Bear} 
\\
        
        \cmidrule(lr){2-4}
        \cmidrule(lr){5-7} 
        \cmidrule(lr){8-10} 
 
        & IoU$\uparrow$ & ATE$\downarrow$ & RMSE$\downarrow$ 
        & IoU$\uparrow$ & ATE$\downarrow$ & RMSE$\downarrow$ 
        & IoU$\uparrow$ & ATE$\downarrow$ & RMSE$\downarrow$ 
        \\
        
        \midrule

        CFGS*
     
        &0.079 & 0.392 & 0.425
        &0.252  & 0.425  &  0.458
        &0.199 & 0.319 &  0.364 
         \\

        4DGS

        & 0.144  & 0.553  & 0.592
        & 0.141  & 0.505  & 0.556
         &Failed   &Failed  & Failed 
         \\
        
        Mo-GS

        & 0.299 &0.267  &0.313
        & 0.298 &0.557  & 0.629 
         & 0.179  & 0.526  & 0.573
         \\

        Dy-GS 
       
        & 0.262  & 0.439  &  0.470 
        & 0.679  & 0.094  &   0.155 
        & Failed   & Failed   & Failed 
         \\
        
        PMGS

        & \textbf{0.940} & \textbf{0.055}  & \textbf{0.086}
        & \textbf{0.876} & \textbf{0.021}  & \textbf{0.093}
        & \textbf{0.969} & \textbf{0.043}  & \textbf{0.047}

        \\
        
        \bottomrule
    \end{tabular}
    }
   \caption{Comparison of motion recovery (Real data).}
       \label{5}
\end{table}

\begin{table}[t]
    \centering
    \renewcommand{\arraystretch}{1}

    \label{ablation}
    \resizebox{\columnwidth}{!}{
    \begin{tabular}{l | c c c | c c c }
        \toprule
        & \multicolumn{3}{c|}{\textbf{Video reconstruction}} 
        & \multicolumn{3}{c}{\textbf{Motion recovery}} \\
        
        \cmidrule(lr){2-4} \cmidrule(lr){5-7}
        
        & PSNR$\uparrow$ & SSIM$\uparrow$ & LPIPS$\downarrow$ & IoU$\uparrow$ & ATE$\downarrow$ & RMSE$\downarrow$ \\
        \midrule
        
        CFGS*   & 16.82 & 0.799 &  0.249  &0.206  & 0.426 & 0.478 \\
        
        Model 1  & 24.61 & 0.910 & 0.089 & 0.635 & 0.334 & 0.791 \\
        
        Model 2 &25.63  &  0.929 &  0.045 & 0.886  & 0.137  & 0.188  \\
        
        Model 3  &28.95 &  0.940& 0.035 & 0.933  & 0.098  &  0.147   \\

        Full & \textbf{29.35} & \textbf{0.949} & \textbf{0.029} & \textbf{0.967} & \textbf{0.080}  &  \textbf{0.133} \\

        \bottomrule
    \end{tabular}
    }
        \caption{Ablation study of different models.}
        \label{Table3}
\end{table}
\noindent
\textbf{Step3-Correction:} In the update step, we use observations (optical flow and learned displacement) to correct the predicted state. These two measurements provide direct observations of displacement, but each carries distinct noise: (1) $z_{flow}^{t}$ is computed from optical flow (Eq.9), representing apparent motion-based displacement observation, with noise denoted as $\sigma_{flow}^{2}$. (2) $z_{learn}^{t}$ derives from the optimization process (Eq. 11), based on Gaussian rendering and loss functions, with noise denoted as $\sigma_{learn}^{2}$. Thus, we define the observation equation as follows:
\begin{equation}
\begin{aligned}
& \textbf{z}_{t} = \begin{bmatrix}
  z_{flow}^{t} \\
  z_{learn}^{t}
\end{bmatrix}= \textbf{H}\textbf{X}_{t|t} + \textbf{v}_{k}, \\
 & \textbf{H} = \begin{bmatrix}
  1&0 \\
  1&0 
  \end{bmatrix}, \textbf{R} = \begin{bmatrix}
   \sigma_{flow}^{2} & 0\\
  0 & \sigma_{learn}^{2} \end{bmatrix}
\label{z}
\end{aligned}
\end{equation}
where $\textbf{H}$ is the observation matrix, $\textbf{R} = \mathbb{E}[\textbf{v}_{k}\textbf{v}_{k}^{T}] $ is the covariance of noise $\textbf{v}_{k}$.

\noindent
\textbf{Step4-Update:}
Based on the above content, we update the Kalman gain $\textbf{K}_{t}$:
\begin{equation}
\textbf{K}_{t} = \textbf{P}_{t|t-1}\textbf{H}^{T}(\textbf{H}\textbf{P}_{t|t-1}\textbf{H}^{T}+\textbf{R})^{-1}
\end{equation}
Finally, we can obtain the optimal estimated value at the current timestamp:
\begin{equation}
\textbf{X}_{t|t} = \textbf{X}_{t|t-1}+ \textbf{K}_{t}(\textbf{z}_{t}-\textbf{H}\textbf{X}_{t|t-1})
\end{equation}

Throughout the fusion process, we fully integrate the physical model, optical flow observations, and network outputs. The Kalman gain is used to automatically adjust the weights, thereby minimizing the error accumulation across multi-modal observation sources.

\section{Experiments}
\subsection{Experimental Settings}

\textbf{Datasets.} 
We constructed both synthetic and real-world datasets. For synthetic data, we collected 6 models from public 3D communities \cite{Blender}, then set up a constant gravity field in Blender. The objects were launched with a randomly set initial velocity and accompanied by autorotation. A fixed camera was used to capture the dynamic scene at 120 FPS, rendering 120 images at a resolution of 1024x1024. For the real-world dataset, we threw 3 different objects and captured the scene using a fixed camera, with an exposure time of 1/2500s, a frame rate of 60 FPS, and a resolution of 4128×2752. During training, the images were resized to below 1600 pixels. Due to the frame rate limitations of the imaging device and the real-world projectile objects do not undergo sufficiently complete rotations (i.e., failing to exhibit a full 360° surface), we captured three projectile sequences for each model from the same viewpoint. Samples were then randomly selected from these sequences to perform modeling. 

For reconstruction, the training/validation split strictly adheres to 3DGS. As for the motion recovery, we focus on recovering the full-sequence motion (all frames), thus no split is required. All competitors follow the aforementioned dataset division for fair metric calculation.

Furthermore, we conducted a comprehensive comparison between our proposed dataset and existing 4D datasets, as detailed in Table \ref{datsests}. PMGS covers both real and synthetic sources and sets the challenging monocular scenario, focusing on long-duration, large-span complex rigid motions. All motions strictly adhere to real-world physical laws, which has not been considered in many synthetic datasets.

\noindent
\textbf{Metrics.}
We evaluate video reconstruction and motion recovery separately. For \textbf{video reconstruction}, we use PSNR, SSIM, and LPIPS \cite{SSIM, PSNR}. For \textbf{motion recovery}, we compute the bounding boxes of the target object in both the real and rendered image, and then calculate the IoU, absolute trajectory error (ATE) and RMSE \cite{IoU, Tracking, CF} to comprehensively assess the spatial accuracy of tracking and trajectory. All evaluations are computed after fully removing backgrounds.

\begin{figure}[t]
\centering
\includegraphics[width=\linewidth]{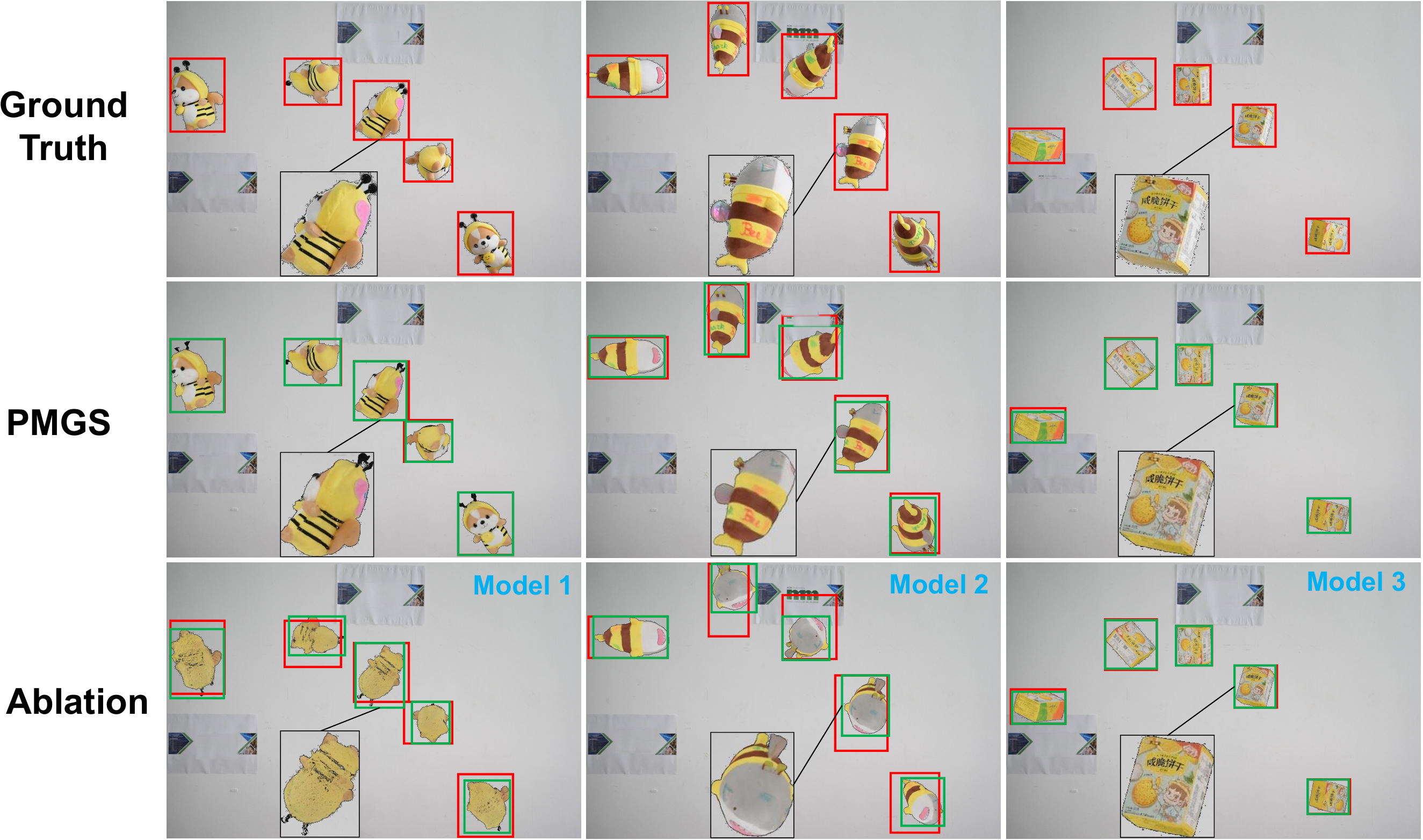}
\caption{Ablation results for different comparison models. PMGS in its complete form exhibits superior stability.}
\label{ablationfigure}
\end{figure}

\noindent
\textbf{Implementation Details.} We conducted experiments in the PyTorch framework with an Nvidia RTX 4090 GPU. During the centralization phase, we cropped the target from the original data and placed it at the center of a 512x512 canvas. In the motion recovery phase, the initial base learning rate applied to the frame with the minimal velocity was set to 1.0e-3, with a base iteration count of 1000. The weights for the $\mathcal{L}_{GS}$, $\mathcal{L}_{Acc}$ and $\mathcal{L}_{Smooth}$ were set to 0.7, 0.2 and 0.1.

\subsection{Comparison}
We select 2DGS \cite{2DGS}, InstantSplat \cite{Instantsplat} as \textbf{static competitors}. \textbf{Dynamic competitors} include 4DGS \cite{4DGS}, DynamicGS \cite{Dynamic3DGS} and MotionGS \cite{MotionGS}. Furthermore, we employ CFGS \cite{CF} as a \textbf{baseline} by disabling its depth estimation for reconstruction, and instead inputting our pre-trained Gaussian model for purely 6DoF pose estimation comparison (symbolized as CFGS*).

\noindent
\textbf{Comparison of video reconstruction.} 
We report the quantitative results in Tables \ref{2} and \ref{3}. Our algorithm demonstrates robust performance, as the proposed density control strategy effectively enhances appearance and geometric quality, thereby providing a solid foundation for dynamic reconstruction. CFGS underperforms, primarily due to inherent flaws in depth inference pipelines. For object-specific modeling under non-open scenarios, the monocular observation system yields an underdetermined solution manifold, where maintaining 3D consistency over extended spatiotemporal intervals becomes theoretically unattainable. 

\noindent
\textbf{Comparison of motion recovery.}  
We uniformly sample frames from complete sequences and visualize in Figure \ref{comparison}, and the quantitative results are reported in Tables \ref{4} and \ref{5}. PMGS performs well in all three tracking metrics, demonstrating accurate estimation of both translation and rotation. 4DGS and MotionGS exhibit obvious trajectory fractures and artifacts—typical limitations of dynamic algorithms when modeling large-scale motions. While DynamicGS demonstrates promising performance, it exhibits limited generalization capability to certain instances. For CFGS, even with our well-trained 3D model provided, the spatial position is incorrectly driven toward infinity under only photometric supervision to forcibly fit image similarity. These findings collectively reveal the effectiveness of physics constraints for recovering complex motions across large spatiotemporal domains.

\subsection{Ablation Study}
We conduct ablation experiments to verify the effectiveness of different novel modules.

\noindent
\textbf{Effectiveness of point density control.} As evidenced in Table \ref{Table3}, Model 1 exhibits degradation, demonstrating that poor reconstruction directly compromises tracking accuracy. Incorrect geometry and appearance representations may cause target mislocalization—especially when floating Gaussian points distort the calculation of central points and bounding boxes, thus resulting in cumulative errors over time.

\noindent
\textbf{Effectiveness of physical constraints.}
Model 2 ablates $L_{Acc}$ and the DSA strategy. As shown in Figure \ref{ablationfigure}, the spatial position of the object exhibits drift, and errors are observed in rotational orientation. The removal of DSA significantly increases the optimization time cost, revealing the necessity of adaptive adjustment of the learning rate under high-speed displacement. 

\noindent
\textbf{Effectiveness of Kalman fusion.} The removal reduces computational overhead at the expense of a decrease in stability. In more challenging scenarios (e.g., under substantial time-varying force interference), the strategy’s role in ensuring robustness would be more pronounced.

\section{Conclusion}

In this paper, we propose PMGS for reconstructing projectile motion over large spatiotemporal scale from monocular videos. Through a two-stage pipeline involving dynamic scene decomposition modeling and physically enhanced motion recovery, we effectively alleviate the issues of trajectory fragmentation and physical implausibility encountered by existing dynamic neural rendering methods when handling high-speed, non-linear rigid motion. In the future, we plan to delve into more challenging topics, such as achieving accurate 3D motion reconstruction under low-quality imaging or changing force field environments.

\section{Acknowledgements}

This work was supported in part by the National Natural Science Foundation of China under Grant 82571371 and in part by the Joint Fund Project of the Ministry of Education for Equipment Pre-research under Grant 8091B042238 (\textit{Corresponding author: Chu He.}), and the Fundamental Research Funds for the Central Universities Grant 204205kf0063 and the National Natural Science Foundation of China under Grant 62271354 (\textit{Corresponding author: Lei Yu.}).

\bibliography{aaai2026}

\end{document}